\documentclass[11pt, a4paper]{article}

\usepackage[margin=1in]{geometry} 
\usepackage{setspace}
\usepackage[utf8]{inputenc}
\usepackage[T1]{fontenc}
\usepackage{microtype} 
\usepackage{libertine} 
\usepackage[libertine]{newtxmath} 
\usepackage{authblk}

\usepackage{graphicx}
\usepackage{amsmath}

\usepackage{booktabs}
\usepackage{array}
\usepackage[skip=10pt]{parskip}
\usepackage{caption}
\usepackage{titlesec}
\titleformat{\section}{\Large\bfseries\sffamily}{\thesection}{1em}{}
\titleformat{\subsection}{\large\bfseries\sffamily}{\thesubsection}{1em}{}

\usepackage{hyperref}
\usepackage{apacite}
\usepackage{tikz}
\usetikzlibrary{arrows.meta, positioning, shapes.geometric, fit, backgrounds}
\usepackage{xcolor}
\usepackage{float}
\usepackage{fontawesome5}
\newcommand{\reslink}[3]{\href{#2}{#1\ \textbf{#3}}}
\hypersetup{
    colorlinks=true,
    linkcolor=blue!70!black, 
    citecolor=blue!70!black, 
    urlcolor=blue!70!black,  
    pdftitle={SHARE: Social-Humanities AI for Research and Education},
    pdfauthor={João Gonçalves, et al.}
}

\title{\textbf{\textsf{SHARE: Social-Humanities AI for Research and Education}}}

\begin{document}

\author[1]{João Gonçalves\thanks{Contact: \href{mailto:ferreiragoncalves@eshcc.eur.nl}{ferreiragoncalves@eshcc.eur.nl}}}
\author[1]{Sonia de Jager}
\author[2]{Petr Knoth}
\author[2]{David Pride}
\author[1]{Nick Jelicic}

\affil[1]{\textit{Erasmus University Rotterdam, NL}}
\affil[2]{\textit{Open University, UK}\protect\vspace{18pt}}

\affil[ ]{\normalsize
\reslink{\faGithub}{https://github.com/Joaoffg/SHARE}{Code}\quad$\vert$\quad
\reslink{\faDatabase}{https://huggingface.co/datasets/Joaoffg/Cloze-SSH}{Dataset}\quad$\vert$\quad
\reslink{\faCube}{https://huggingface.co/Joaoffg/SHARE-14B-Base-2604}{SHARE-14B}\quad$\vert$\quad
\reslink{\faCube}{https://huggingface.co/Joaoffg/SHARE-4B-Base-2604}{SHARE-4B}
}

\maketitle

\begin{abstract}
This intermediate technical report introduces the SHARE family of base models and the MIRROR user interface. The SHARE models are the first causal language models fully pretrained by and for the social sciences and humanities (SSH). Their performance in modelling SSH texts is close to that of general purpose models (Phi-4) which use 100 times more tokens, as shown by our custom SSH Cloze benchmark. The MIRROR user interface is designed for reviewing text inputs from the SSH disciplines while preserving critical engagement. By prototyping a generative AI interface that does not generate any text, we propose a way to harness the capabilities of the SHARE models without compromising the integrity of SSH principles and norms.

\end{abstract}

\section{Introduction}

Large language models (LLMs) are increasingly being used to accelerate scientific research. However, their adoption, particularly for the production of scientific outputs such as papers, reports, and grant proposals, has been accompanied by concerns related to quality, bias \cite{grossmann2023ai}, and an erasure of critical thinking \cite{liu2025generative}. These concerns are particularly pressing for social sciences and humanities (SSH) domains, where open-ended exploration and imagination are at the core of academic research and education.

While the worries raised in relation to LLMs are valid, they are also rooted in a way of engaging with models that is constrained to the chat-like interaction offered by commercial LLM providers, one which is strongly output- (rather than analysis-) oriented. As such, it is unclear if risks for social scientific work associated with LLM usage are driven by the technology itself or the way it is deployed by large commercial companies. We therefore ask, how can LLM technology be developed and deployed in a way that enhances the critical thinking of SSH domains?

We address the first part of this question, development, by proposing the \textit{Social-Humanities AI for Research and Education} (SHARE) models, a family of transformer-based LLMs pretrained exclusively on content relevant to SSH scholarship. We then treat the second part with the \textit{Model Interface for Reflective Research Output Revisions} (MIRROR), a user interface for SHARE that is inspired by LLM plagiarism-detection but that aims instead to signal novelty and discovery through the exploration of unexpectedness in research outputs.

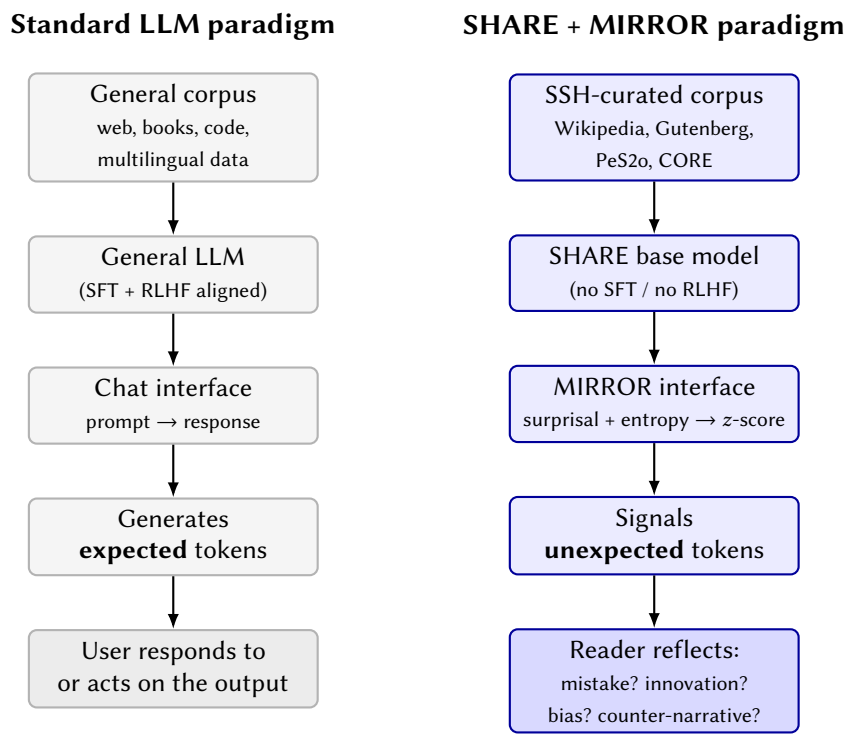
\begin{figure}[htbp]
\centering
\begin{tikzpicture}[
    font=\sffamily\small,
    node distance=0.7cm and 1.2cm,
    box/.style={
        rectangle, rounded corners=3pt,
        draw, thick,
        minimum width=3.8cm, minimum height=1cm,
        align=center,
        inner sep=4pt
    },
    stdbox/.style={box, draw=gray!60, fill=gray!8},
    sharebox/.style={box, draw=blue!60!black, fill=blue!8},
    arr/.style={-{Latex[length=2mm]}, thick},
    panellabel/.style={font=\sffamily\bfseries}
]

\node[stdbox] (gencorpus) {General corpus\\ \scriptsize web, books, code,\\ \scriptsize multilingual data};
\node[stdbox, below=of gencorpus] (genmodel) {General LLM\\ \scriptsize (SFT + RLHF aligned)};
\node[stdbox, below=of genmodel] (chat) {Chat interface\\ \scriptsize prompt $\rightarrow$ response};
\node[stdbox, below=of chat] (genexp) {Generates\\ \textbf{expected} tokens};
\node[stdbox, below=of genexp, fill=gray!15] (output) {User responds to\\ or acts on the output};

\draw[arr] (gencorpus) -- (genmodel);
\draw[arr] (genmodel) -- (chat);
\draw[arr] (chat) -- (genexp);
\draw[arr] (genexp) -- (output);

\node[panellabel, above=0.3cm of gencorpus] {Standard LLM paradigm};

\node[sharebox, right=2.5cm of gencorpus] (corpus) {SSH-curated corpus\\ \scriptsize Wikipedia, Gutenberg,\\ \scriptsize PeS2o, CORE};
\node[sharebox, below=of corpus] (share) {SHARE base model\\ \scriptsize (no SFT / no RLHF)};
\node[sharebox, below=of share] (mirror) {MIRROR interface\\ \scriptsize surprisal + entropy $\rightarrow$ $z$-score};
\node[sharebox, below=of mirror] (signal) {Signals\\ \textbf{unexpected} tokens};
\node[sharebox, below=of signal, fill=blue!15] (reader) {Reader reflects:\\ \scriptsize mistake? innovation?\\ \scriptsize bias? counter-narrative?};

\draw[arr] (corpus) -- (share);
\draw[arr] (share) -- (mirror);
\draw[arr] (mirror) -- (signal);
\draw[arr] (signal) -- (reader);

\node[panellabel, above=0.3cm of corpus] {SHARE + MIRROR paradigm};

\end{tikzpicture}
\caption{Two pipelines for LLM use in scholarly work. The standard pipeline generates expected text from a general-purpose aligned model; the SHARE + MIRROR pipeline uses an SSH-specialised base model to signal unexpected tokens for the reader to reflect.}
\label{fig:conceptual_pivot}
\end{figure}

Taken together, these innovations offer a new avenue for LLM use in SSH. By offering a method for self-reflection in relation to the English SSH literature,\footnote{In the context of this project we are bound primarily to materials in English, and to a lower extent in Dutch, but we consider this a limitation and would like to explore multilingual aspects in the future.} the models can be used to treat and modify aspects of academic writing, but also to identify different disciplinary biases and---style, domain, provenance, vocabulary, etc.---constraints by signalling deviations from what would be the expected outputs. This approach is summarized in Figure \ref{fig:conceptual_pivot}.

This intermediate technical report, released as SHARE-14B has completed 15\% of training, aims to share the progress and results we have achieved so far in the training of the SHARE models to collect feedback and suggestions for the remaining stages of training and development.

\section{Related work}

LLMs are Natural Language Processing (NLP) algorithms designed to tackle challenges related to the complexity of modelling language structures through high parameter counts. Causal language models, which aim to predict the next token\footnote{Tokens are the units by which strings of characters are chunked in order to become amenable to analysis and recombination. One can immediately imagine how issues of linguistic; cultural; disciplinary; etc., consideration become crucial in design and implementation here: which alphabets and for what purposes are we considering? What ought to be considered “basic” units of (a) language? And are these always the same for all domains and in all languages? The answers to these and related questions are much more complex than assumed in most implemented LLMs to date.} in a text, have become a popular tool to support a wide range of tasks, driven by their accessibility through commercial providers such as OpenAI \cite{singh2025openai}, Google DeepMind and Anthropic. Recently, scientific discovery has been one of the key claims from model providers in relation to the potential of LLMs \cite{openai2026prism}.

However, most of the developments and claims in relation to AI impact on scientific work do not come from SSH domains, but from fields such as mathematics and coding \cite{guo2025deepseek}, engineering \cite{jiang2025autotriz}, or biochemistry \cite{zhang2025scientific}. To some extent, this focus might be justified by the technical developments, such as Group Relative Policy Optimization (GRPO), which are more efficient in domains that have verifiable rewards, for which the interpretative nature of SSH is less suited. Complementarily, these priorities mimic a systemic imbalance in attention, funding and resources \cite{kyriakidis2025focus} where SSH domains are often framed as secondary, especially in relation to  innovations associated to STEM field such as LLMs, because their economic value is not as immediately accessible.\footnote{Interesting to note is, however, how the general discourse pertaining to the possible purposes and uses of so-called “Generative artificial intelligence” is veering towards creativity and open-endedness as typically explored through the humanities, challenging traditional “efficiency” narratives.}

Some efforts to train or use LLMs for the social sciences and humanities have been made. SSciBERT is a masked LLM pretrained exclusively for SSH abstracts \cite{shen2023sscibert}, aiming to excel at tasks such as discipline classification and named-entity recognition within social science texts. Other models, such as BERT-NLI \cite{laurer2024less}, are not specific to the SSH but have been developed with SSH use cases largely in mind, such as delivering high performance text classification of political content at reduced costs. However, to our knowledge, there are no efforts at training auto-regressive LLMs (often labelled Generative AI) with the number of tokens or parameters required to achieve minimal performance, even for a domain-specific language model. The SHARE family of models fill this gap, by first providing 4 billion and 14 billion parameter models pretrained on more than tens of billions of tokens relevant to the SSH.

\section{The role of expectations}

Expectations play a key role in both the development of SHARE and its envisioned deployment. Over the course of the past few years, ChatGPT's chat interface and template \cite{ouyang2022training} have become the \textit{de facto} standard for how individuals, including social scientists and humanities scholars, expect to interact with LLMs.\footnote{We see this trend and its implicit expectations as following from users having become accustomed to comparable interfaces in earlier technologies: from texting on mobile phones to message boards and social media. On the one hand this may be considered neutral and innocuous: this dialogical interaction model can be understood as straightforward, dating back to, e.g., Turing's imitation game. However, it can also be understood as problematic when its ELIZA effect renders credibility and a sense of trust in something which is in fact rather volatile and lacking in the consistency humans normally expect from interactions with other human interlocutors.} In fact, during earlier work on the Erasmian Language Model (ELM) \cite{gonccalves2024advantages}, a university model pretrained exclusively on data from Erasmus University Rotterdam, comparisons with ChatGPT were inevitably a feature in most formal and informal interactions. This means that the expectation for causal LLMs is that their primary usefulness stems from generating text from a probability distribution that selects among the most likely tokens that should follow the user input. This is a particularly relevant consideration for the SSH domains, where academic work requires the exploration of previously disconnected variables and concepts, the re-framing and reconfiguration of these very concepts, and the questioning of power structures\footnote{Anything from historical narratives; dominant ideologies; established semantic frameworks; etc., all the way to the very structures of language itself, as mentioned earlier.} that also shape language use itself, including within academic work of SSH. While methodological and theoretical rigour are cornerstones of the Social Sciences and Humanities, predictability weakens the very foundations of the field by constraining thought and exploration.\footnote{And while the vast combinatorial possibilities afforded by LLM use for exploration in SSH are resourceful, it does seem as if the leap required to make relevant novel connections still remains within the exclusively human-capacity realm, as we are often dealing with things that are relevant to highly specific human needs and desires (semantic saliences; aesthetic trends; political sentiments; historical consciousness, etc.).}

What is then the use of an autoregressive LLM for the core of SSH work? We argue that, for a model pretrained exclusively on an SSH corpus, predicting the next token is equivalent to making explicit what is expected in these disciplines.\footnote{\textit{Expected} in the sense that it is accepted, tacit and normalized, less in the sense that it is “accurate” or methodologically established, which would be the case we referred to earlier in, e.g., STEM disciplines.} This means that, rather than being used to generate expected and predictable texts, the model can be used to assess the extent to which texts produced in the scope of the SSH conform to or deviate from the expectations of the field. We believe that by \textit{signalling} the unexpected instead of \textit{generating} the expected, SHARE can become a tool for disciplinary (meta-)reflection and exploration rather than unnecessarily overproductive automation, which can be understood as negatively impacting the analytical imperative and creative, selective drive of SSH disciplines.

In our treatment of expectations in SSH writing, we apply expectancy violation theory (EVT) \cite{burgoon2015expectancy}, which makes “the counterintuitive claim that violations of expectations are sometimes preferable to confirmations of expectations".\footnote{Precisely, novelty as \textit{deviation from expectation} is the hallmark of---literary; artistic; aesthetic; and related types of what we tend to call---\textit{creativity}, as explored in computational and cognitive neuroscience literature on the subject \cite{clark2017nice}. The disciplines which analyze and determine trends, movements, etc., by studying what kinds of novelty become culturally salient; relevant and to which discourses, are precisely SSH domains, which is why we find SHARE, and its interface, a highly pertinent project in the context of LLMs for science.} While this theory originated from interpersonal communication, we believe that the same principle applies to scholarly work, at least within the SSH domains. In written scientific language, deviations from expectations can sometimes signal undesired aspects, such as a typo in a text; the misuse of a concept; or lack of attribution of scholarly work. These are deviations that most academics would, in principle, agree are negative and should be corrected in a text. They are often most prevalent among those that are newly exposed to the norms and expectations of the field, such as first year bachelor students in an academic skills course, and making these mistakes and correcting them is part of the learning process of becoming a more eloquent scholar.

At the same time, unexpectedness is also signalled as one of the key characteristics of successful scientific writing \cite{schimel2012writing}. Presenting a concept under a new theoretical lens, or proposing a theory that makes a counter-intuitive claim, such as EVT itself, is acknowledged in the SSH as one of the most insightful contributions in a scientific output. From a language modelling perspective, when manifested as tokens, these ideas might have a high perplexity and are seen as deviations from expectation. For an SSH researcher, often, exceptions to the norm are the reason we keep reading and writing papers and books. Additionally, the value of deviation from expectations goes beyond scientific innovation, as it also allows the field and its norms to evolve and correct systemic imbalances that have historically conditioned it. Countries, themes, epistemologies and scholars that have been traditionally ignored or even disregarded by the field due to, e.g., colonial; discriminative; ideological, etc., influences will be flagged as unexpected by a model trained on legacy texts from the field. This shift from `silent to salient' in academic texts is a necessary condition to counter the harmful consequences of these influences. Finally, even when considering that academic writing tends to be a highly standardized practice, the variation, and unexpectedness, that comes with personal writing styles can often be seen to enhance, rather than detract, from reading experiences.\footnote{As it has also become apparent through the massive exposure to commercial LLM outputs in recent years: many stylistic aspects of LLM-text (what we could rightly call “LLMisms”---such as the use of the m-dash, “it's not just A, it's B” type phrases, or bullet-point listing) are already proving to be rather tiresome to encounter, and have increasingly become “old news” for most LLM-sensitive readers.}

We believe that one of the contributions of SHARE is to signal violations of expectations in academic texts under the assumption that, by becoming aware of them, individual authors and the field as a whole can gain from reflecting on them, and their positive or negative contributions to a text. We therefore develop the Model Interface for Reflective Research Output Revisions (MIRROR) user interface, which uses colour coded labels to signal unexpected tokens in user texts, as the default way of interacting with the SHARE base models.

\section{The SHARE Family of Models}

\subsection{Data}
We construct an SSH-focused dataset from three corpora: Wikipedia articles, Project Gutenberg books, and academic publications. While the majority of academic publications were sourced from the PeS2o \cite{peS2o} and CORE \cite{knoth2012core} datasets, a smaller subset was obtained through dialogue and agreement with publishers and authors, such as the Open Humanities Press. A critical limitation of our data is its over-reliance on primarily English datasets which is problematic in SSH fields. However, this was a necessary trade-off considering the resources we had at our disposal for this project, and something we hope to address in future iterations of SHARE.

To identify the Social Sciences and Humanities (SSH) subset within each corpus, we adopt a hybrid approach combining metadata-based heuristics and machine learning classification. When metadata is available, we apply heuristic based filters on features such as categories to select SSH-relevant documents. In cases where this kind of metadata is not available, we use AllenAI’s Field of Science (FoS) classifier to assign disciplinary labels. 

\subsubsection{Wikipedia}
Wikipedia is a free online encyclopedia, organized through a hierarchical category system. It has main topic classifications that are used throughout Wikipedia to organize the presentation of links to articles in its various reference systems (\href{https://en.wikipedia.org/wiki/Category:Main_topic_classifications}{Main topic classifications}). We first select relevant main topic classifications corresponding to SSH disciplines. In Table \ref{tab:wikipedia_cats} below, we present an overview of the included main topics used to identify SSH-related articles.  

Using the PetScan tool, we traverse the category tree from these entry points to collect the identifiers of all associated articles. We then download the English and Dutch Wikipedia dumps and extract the textual content of the selected articles using WikiExtractor (\href{https://github.com/attardi/wikiextractor}{WikiExtractor}). 

\begin{table}[htbp]
    \centering
    \caption{Main Wikipedia Topics}
    \label{tab:wikipedia_cats}
    \vspace{0.2cm}
    \begin{tabular}{ll}
        \toprule
        \textbf{Main topic} & \textbf{Notes} \\
        \midrule
        Business       & \\
        Communication  & Excluding animal and plant communication \\
        Culture        & \\
        Economy        & \\
        Education      & \\
        Geography      & \\
        Government     & \\
        History        & \\
        Human behavior & \\
        Humanities     & \\
        Language       & \\
        Law            & \\
        Philosophy     & \\
        Politics       & \\
        Religion       & \\
        Science        & Only Social Sciences and Formal Sciences \\
        Society        & \\
        \bottomrule
    \end{tabular}
\end{table}

\subsubsection{Project Gutenberg}

Project Gutenberg is a collection of historical books, primarily consisting of public domain works. The corpus provides broad coverage of literary and scholarly texts relevant to SSH domains.

We identify relevant Library of Congress Classes (LOCC) corresponding to SSH disciplines using the available metadata. Based on these categories, we select and download the corresponding subset of books. The full text of the selected works is then extracted for further processing.

Table \ref{tab:gutenberg_cats} below shows a list of the included LOCC used to identify SSH related books in Project Gutenberg.

\begin{table}[H]
    \centering
    \caption{Gutenberg Categories}
    \label{tab:gutenberg_cats}
    \vspace{0.2cm}
    \begin{tabular}{ll}
        \toprule
        \textbf{Category} & \textbf{Name} \\
        \midrule
        Class B & Philosophy, Psychology, Religion \\
        Class C & History \\
        Class D & History: General and Eastern Hemisphere \\
        Class G & Geography, Anthropology, Recreation \\
        Class H & Social sciences \\
        Class J & Political science \\
        Class K & Law in general, Comparative and uniform law, Jurisprudence \\
        Class L & Education \\
        Class M & Music \\
        Class N & Fine Arts \\
        \bottomrule
    \end{tabular}
\end{table}

\subsubsection{PeS2o and CORE}

The peS2o dataset is a collection of around 40 million open-access academic papers. PeS2o is derived from the Semantic Scholar Open Research Corpus \cite{lo2020s2orc}.

CORE (COnnecting REpositories) operates as a large-scale aggregation infrastructure that systematically harvests, normalizes, and enriches scholarly content from a globally distributed network of institutional and subject repositories. Using standard interoperability protocols such as OAI-PMH, CORE ingests metadata and, where available, full-text content from the global network of over 14,500 repositories worldwide, alongside complementary sources including subject repositories (e.g. arXiv), biomedical archives (e.g. PubMed Central), and other open-access and preprint platforms. As of May 2025, CORE currently hosts some 49.2m full text scholarly documents and 449m metadata records \cite{knoth2023core}.

A key component of CORE’s contribution lies in its large-scale text acquisition and processing capabilities. Full texts are deduplicated, parsed, and enriched with additional metadata such as DOIs and other unique identifiers. This results in a high-coverage, research-grade dataset that is a comprehensive record of global open-access scholarship. CORE’s role is therefore not only infrastructural but epistemic: by aggregating and standardizing literature at scale, CORE enables the construction of domain-specific datasets that underpin the SHARE models and supports the empirical analyses presented in this work.

To identify SSH-relevant academic publications within peS2o we directly use the available Field of Science (FoS) labels from the AllenAI classifier to identify relevant works. For CORE, where these labels are not available, we create FoS labels by applying the same classifier to titles, abstracts, and where available publication venues (\url{https://github.com/allenai/s2_fos}).

We then retain publications assigned to the relevant SSH fields to construct the final subset. Table \ref{tab:fos_cats} below shows the relevant fields used.

\begin{table}[H]
    \centering
    \caption{Fields of Science Categories}
    \label{tab:fos_cats}
    \vspace{0.2cm}
    \begin{tabular}{l}
        \toprule
        \textbf{Field of Science} \\
        \midrule
        Art \\
        Business \\
        Economics \\
        Geography \\
        Education \\
        History \\
        Law \\
        Linguistics \\
        Philosophy \\
        Political Science \\
        Psychology \\
        Sociology \\
        \bottomrule
    \end{tabular}
\end{table}

\subsection{Model architecture}

The SHARE models are decoder-only transformer language models based on Microsoft's Phi-4 architectures \cite{abdin2024phi}. Like Phi-4, the baseline context length was 4096 tokens, which was expanded to 8192 tokens after the initial pretraining. Instead of the default phi-4 tiktoken tokenizer, we trained a custom byte-pair encoding (BPE) tokenizer on our full training corpus for SHARE-4B. This tokenizer has a reduced vocabulary size of 50 000 tokens when compared to Phi-4's 100 352 tokens. This is because, unlike the multilingual phi-4, our data is mostly in English and Dutch, requiring a less extensive vocabulary to be effective, even considering highly specialized academic texts. Inspection of tokenization before pretraining showed the tokenizer to be suitable for academic texts, breaking down words in reasonable and understandable locations and keeping frequently used words whole. Similarly to Phi-4, we trained two versions of the model, one with 3.9 billion parameters (SHARE-4B) and one with 14 billion (SHARE-14B), they mimic, respectively, the Phi-4-mini and Phi-4 architectures.

\subsection{Training}

For our pretraining runs, we based our approach on examples of fully open pretraining runs, especially of the OLMO 2 models \cite{olmo20242}.

Pretraining of the SHARE-4B model was made possible by an NVIDIA Academic Grant. Training took place in the Saturn Cloud environment using data parallelism across 8 NVIDIA-A100 GPUs over a period of 656 hours to complete 2 epochs of the training data described above, excluding the CORE dataset (total of 28 billion tokens). Given our limited amount of data, an evaluation set of 500 examples was used. A global batch size of 64 was used, a weight decay of .01 and a learning rate of 2e-4 with 3000 warmup steps and a cosine learning rate scheduler. In general, we aimed to replicate successful pretraining recipes of models like OLMO, tweaking hyperparameters if necessary to reflect our shorter training periods and amount of data. Generally, training loss, evaluation loss and gradient normalization values indicate a smooth training run, with loss decreasing gradually until reaching an evaluation perplexity of 11.94 in the evaluation set. While carbon emissions for Saturn Cloud are difficult to estimate, an average assumption would be that emissions tied to training amounted to about 1.2 metric tons of Co2 equivalents, roughly equivalent to the emissions associated with an economy one-way flight ticket from Amsterdam to New York.

Pretraining of the SHARE-14B model is currently under way. So far, we have trained the model on a total of 96 billion tokens\footnote{When migrating from Saturn Cloud to Snellius, we opted to not exclude the data that was already seen during the early stages of pretraining, meaning that the final number of pretraining tokens will be slightly higher than just 2 epochs}. For the final version of the model, we aim to complete 2 epochs of the training data, reaching a compute optimal 630 billion tokens, meaning that 15\% of training has been completed for this technical report. Training of the 14B parameter model was initiated in the Saturn Cloud environment using compute provided by the NVIDIA Academic Grant on 8 80GB A100 GPUs for 167 hours using fully sharded data parallel (FSDP). It was then continued with a Small Compute Grant (EINF-15690) from SURF and NWO, using five (5) nodes of four (4) H100 GPUs each for a total of twenty (20) GPUs running in parallel for approximately 225 hours on the Dutch supercomputer Snellius. Given the large scale of pretraining, we spent considerable effort to ensure the highest possible throughput in the H100 GPUs, applying torch.compile, the Liger Kernel, and packing, in addition to the use of FlashAttention-2 already used in the 4B parameter model. We have started with 2000 warm-up steps reaching a learning rate which is being manually monitored and adjusted after every five (5) day run on Snellius (starting at 1.58e-4 for the first run, adjusted to 1e-4 for the second one). While unusual, we considered manual adjustments to be appropriate due to the specific nature of the training data and because of works cautioning that cosine learning rate decay may underutilize training data fed in the later stages of pretraining \cite{NEURIPS2024_8b970e15}, which is a concern given the filtered nature of our dataset. We have also increased the weight decay to .1 from the 4B parameter model. Like above, we have observed smooth training runs without significant grad norm spikes and a gradual decrease in both training and evaluation loss. After 15\% of training, we observe an evaluation perplexity of 5.26, hinting that the 14B model is already significantly more capable than the 4B parameter model. Both models underwent context expansion to an 8k token context window with 5000 English academic documents extracted from PeS2o.

\section{The MIRROR interface}

For most LLM development projects (see OLMO 3 as an example \cite{olmo2025olmo}), the expected steps after pretraining would be supervised fine-tuning (SFT) and some form of alignment (e.g. Direct Preference Optimization (DPO) or Reinforcement Learning with Human-Feedback (RLHF)), and, optionally, reasoning scaling with GRPO. However, there are no existing SFT, DPO, or RLHF datasets that we are aware of that are specifically constructed for SSH domains. As an example, any attempt to align the model to reject inquiries over topics such as hate speech or racial discrimination would in fact alienate a substantial portion of SSH research on these topics. As such, we elect to make the SHARE models available in their base pretrained form, which we consider to be the most accurate representation of the scientific domains they were trained for. When experimenting with general SFT approaches, we actually saw a degradation of text generation performance, an observation that is supported by recent literature \cite{ye2025analyzing}. In light of this, we shift our efforts towards providing an accessible interface that allows users to extract analytical insights from SHARE without requiring a chat dialogue. In our view, this also has the added benefit of preventing the anthropomorphisation of LLMs, which is associated with some of the key dangers stemming from the use of these models \cite{peter2025benefits}.

Together with the theoretical principles guiding the development of SHARE, MIRROR was guided by three core design principles: 1) Allowing interaction with a model in its pretrained state; 2) Make explicit the expectations associated with texts from the SSH, meaning that outputs should not be prescriptive or normative in nature; 3) Respect for the publishers, authors, and educators that are behind the academic training data for SHARE, minimizing issues such as copyright infringement or (student) plagiarism. Our solution is, therefore, a Generative AI interface that does not generate any text. The baseline process behind MIRROR is to take a user written text as input, and output the surprisal for each token in the text considering the predicted model logits. In other words, it tells the user how unexpected each token is based on its knowledge of the Social Sciences and Humanities, allowing the user to assess by themselves if this unexpectedness is negative (mistake; irrelevance; etc.) or positive (innovation; discovery; etc.).\footnote{See \cite{wu2023predicting} for an implementation of this principle to assess literary quality.}

This form of user interface is inspired by plagiarism detectors and experiments like the Large Language Model Test Room \cite{gehrmann-etal-2019-gltr}. However, while in these cases the purpose is to estimate the likelihood that a text was generated by an LLM, our interface flips this concept by using the LLM to estimate how much a user text is unexpected within the language used by the SSH domains. While this is an unusual proposition, we believe it is particularly suitable for the SHARE model because of its open weights, which allow for logits of each token to be extracted, its domain-specific nature, which ensure that surprisal is measured against a specific type of text, and its lack of further fine-tuning or alignment, which guarantees that these probabilities are computed in what could be called its \textit{purest} form.

The output visualisation for MIRROR aims at presenting surprisal from model outputs. The content and structure of the interface stems from EVT, but its code implementation as a gradio app demo was supported by the Claude Opus 4.6 model. First, a heatmap visualization shows a z-score per token using the full text, accounting for surprisal and entropy of tokens to determine how unexpected they are in relation to the SHARE model predictions. Reversing how a plagiarism detection tool would highlight the sections with the lowest surprisal to show the likelihood that the text would be LLM generated, this visualization makes tokens with high z-score salient so that the reader can assess deviations for SSH expectations. Additionally, when hovering on the token, it presents a thesaurus-like view with the tokens that are most likely to be predicted by the model. However, unlike a thesaurus which presents words similar to the actual token, this function presents expected tokens based on the preceding text, which might actually deviate significantly from the actual token in the text.\footnote{For some foundational considerations what pertains thesauri; machine interpretation; tokenizing and related issues, please consult early NLP scholar Margaret Masterman's outstanding yet often ignored work on the subject \cite{masterman2005language}.}

Surprisal:
$$S_t = -\log p(x_t \mid x_{<t})$$

Entropy:
$$H_t = -\sum_{v \in \mathcal{V}} p(v \mid x_{<t}) \log p(v \mid x_{<t})$$

Standard deviation of surprisal:
$$\sigma_t = \sqrt{\sum_{v \in \mathcal{V}} p(v \mid x_{<t}) \left(-\log p(v \mid x_{<t})\right)^2 - H_t^2}$$

Z-score:
$$Z_t = \frac{S_t - H_t}{\sigma_t}$$

This (1) baseline interface allows for other forms of visualization, such as (2) showing the most unexpected tokens in a text ranked by their surprisal values, (3) ranking tokens that do not appear in the text based on the cumulative probabilities that they should be expected to appear in the text, grouping unexpectedness values per sentence (4) and per paragraph (5), allowing for a higher level overview of conformity or deviations from expectations. 

\section{Results}

Given that the aim of SHARE is to represent the SSH fields, traditional benchmarks normally used to measure LLM performance, such as MMLU, are not a valid test for the model's intended purpose. They assume both content (high school materials, STEM field) and formats (multiple choice tests) that we deliberately excluded from pretraining data. Low scores in these benchmarks would likely stem from a misalignment between the format and content of the measurements, rather than from lack of model capabilities. To address this, we report a set of empirical investigations that aim to evaluate the SHARE models, both in isolation and in relation to comparable models. We first present a quantitative comparison of how SHARE models predict content SSH texts in comparison to the Phi models, we then use a custom Cloze benchmark to show performance in predicting content that requires a high level of SSH contextual understanding, and finally report a qualitative overview of usefulness and shortcomings when used with the MIRROR interface.

\subsection{Relative specialization per scientific field}
\begin{figure}[!htbp]
    \centering
    \includegraphics[width=1\linewidth]{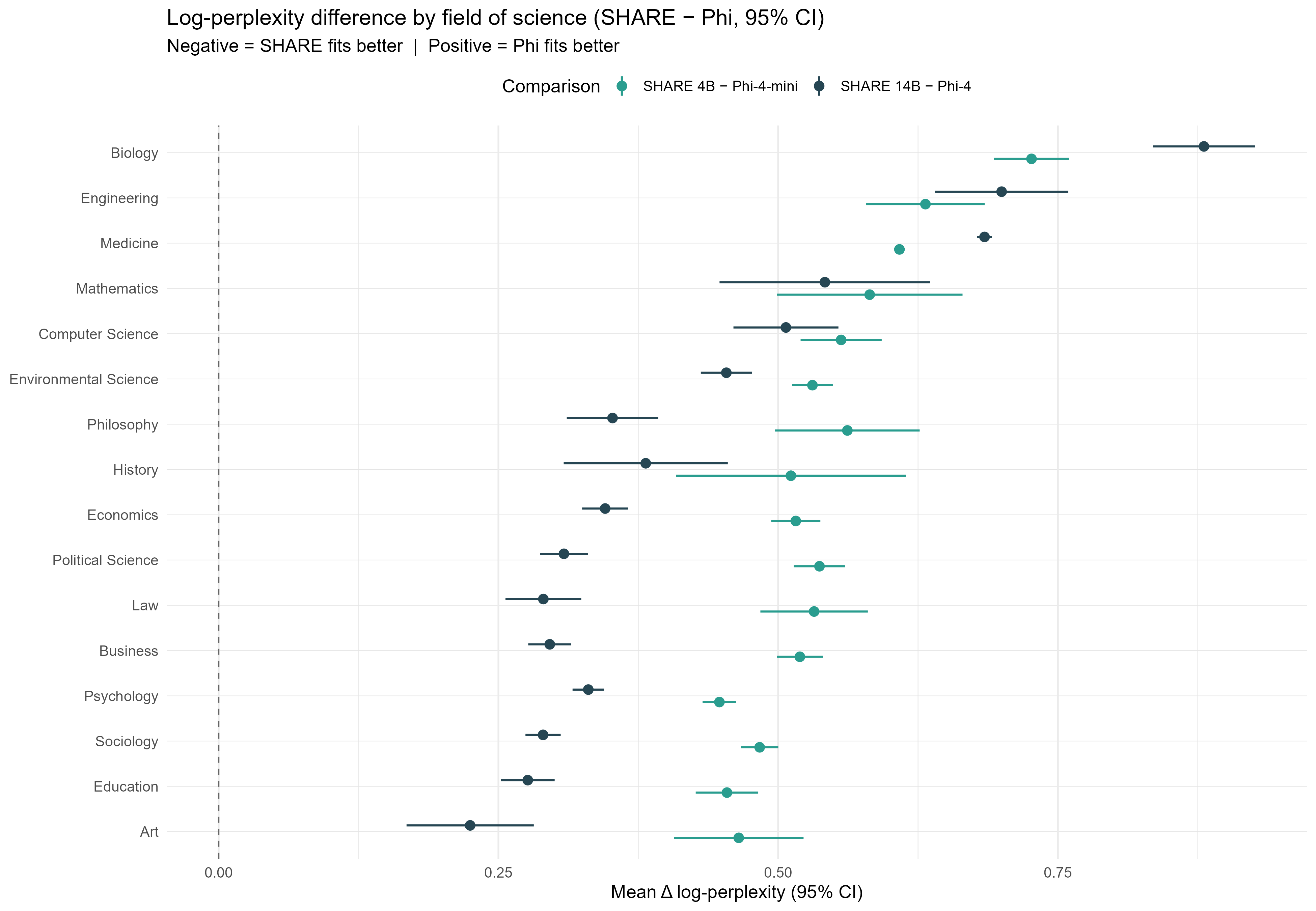}
    \caption{The figure shows the average difference in log perplexity between the SHARE and Phi-4 models per scientific domain following the FoS classifier. Lower values mean that SHARE fits better to that scientific domain in relation to Phi-4. Error bars indicate 95\% confidence intervals.}
    \label{fig:perplexity_field}
\end{figure}

\begin{figure}[!htbp]
    \centering
    \includegraphics[width=1\linewidth]{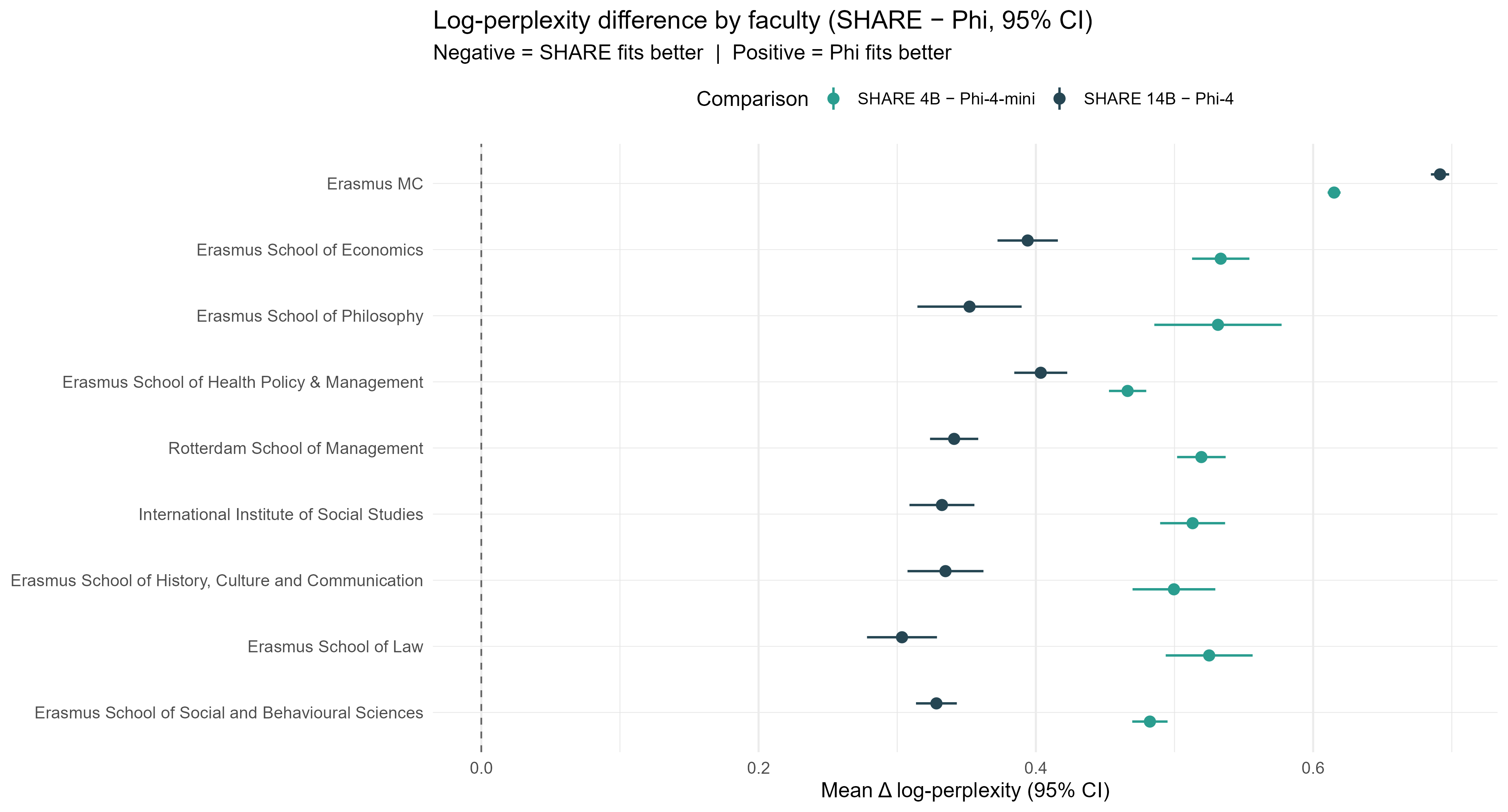}
    \caption{The figure shows the average difference in log perplexity between the SHARE and Phi-4 models based on the faculty that the author is affiliated to at Erasmus University Rotterdam. Lower values mean that SHARE fits better to research outputs of that faculty in relation to Phi-4. Error bars indicate 95\% confidence intervals.}
    \label{fig:perplexity_faculty}
\end{figure}

To validate that the SHARE model is more closely aligned with SSH domains than other scientific disciplines, we compute its perplexity values on a set of abstracts that are not in its training data distribution, namely, research outputs at Erasmus University Rotterdam during the Q3 and Q4 periods of 2025. As a reference point, we compare the average perplexity per scientific domain between the SHARE models and the Phi-4 models (Figure \ref{fig:perplexity_field}), given that the latter are the most similar in architecture to our own models. While we use the same FoS classifier as in our training data to determine the scientific field, which could introduce some endogeneity in validation, we also report results for abstracts based on the faculty affiliation of the authors of the abstract (Figure \ref{fig:perplexity_faculty}). This also acts as an additional check to our FoS text selection, showing that both the model and training data selection are ecologically valid based on faculty affiliations.

A comparison between the SHARE and Phi-4 models supports the specialization of SHARE in SSH domains. In the field of science chart (Figure \ref{fig:perplexity_field}), the log perplexity gap between the two models is consistently smaller for SSH fields such as Art, Education and Sociology and larger for STEM fields such as Biology, Engineering and Medicine. Analysis at a faculty level tells a similar story (Figure \ref{fig:perplexity_faculty}). While Erasmus University is mostly specialized in SSH domains, the medical center (Erasmus MC) offers a useful contrast for comparison, revealing a lower fit with the SHARE models when compared to the other faculties at Erasmus. Taken together, these figures suggest that SHARE shows a more prominent specialization skew for SSH domains than general models such as Phi-4.

It is noteworthy that the Phi-4 models still offer a better fit in general to the abstract texts, shown by the positive values in the figure. However, this could be attributed to the fact that the model is more confident in the English language in general, given that its training data was in the order of the trillions of tokens while the scale of training data for SHARE is in the dozens of billions. We aim to address this limitation with our next test, which focuses on tokens that are more directly relevant for SSH textual processing.

\subsection{The SSH Cloze Benchmark}

To disentangle performance in general English writing from SSH relevant text prediction we created a custom Cloze style \cite{taylor1953cloze} benchmark that aims to predict probabilities of equivalent tokens for which SSH familiarity is key. For instance, in the sentence "The correlation between social media use and well-being was negative.", predicting the word "was" only requires basic knowledge of English sentence structures and verbal tense, which can be acquired with pretraining on general corpora. However, predicting the word "negative" in opposition to "positive", requires domain-specific knowledge from the social sciences. For this benchmark, we sample from and adapt SSH output abstracts to create a token prediction problem where a decision between two equivalent tokens requires specific knowledge on the content and structure of SSH texts.

\begin{figure}[!htbp]
    \centering
    \includegraphics[width=1\linewidth]{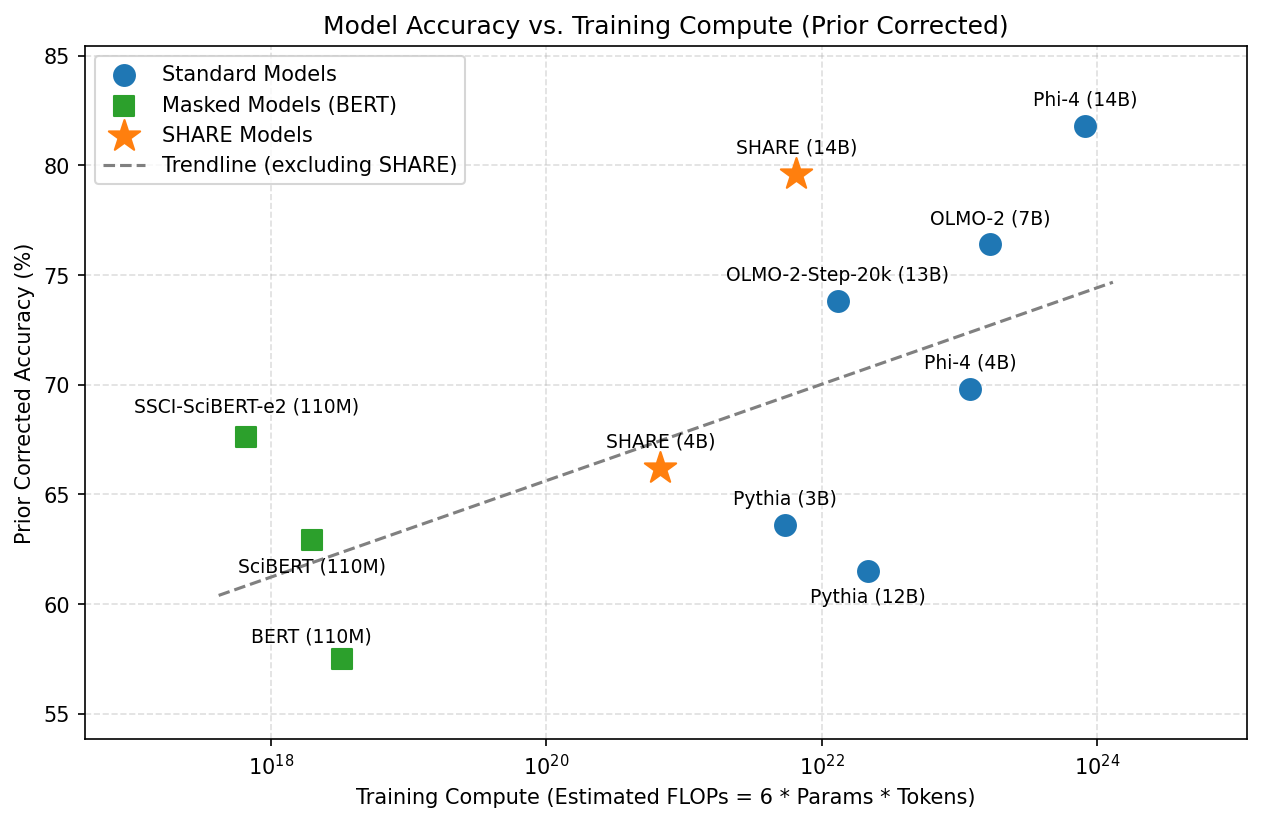}
    \caption{Prior corrected accuracy on the SSH Cloze benchmark (vertical axis) based on estimated flops (horizontal axis, logarithmic scale). Data points in the top left quadrant of the chart indicate more efficient models, achieving better performance with lower training compute.}
    \label{fig:cloze_benchmark}
\end{figure}

We use 275 out of distribution SSH abstracts published in Q1 2026, sampling 25 examples per Web of Science scientific field (Art, Business, Communication, Economics, Education, Geography, History, Law, Philosophy, Psychology, Sociology)\footnote{This is an initial version of the benchmark created for this technical report, but we plan to expand on both the number of examples and disciplines as development continues.}. We use a keyword search aiming to find words (positive OR negative OR lower OR higher OR greater OR smaller) that would lend themselves to equivalent token assessment on Web of Science, ranking results per discipline based on their citation count. Recency was a key requirement to account for the limitations of current benchmarks which have format limitations and possible data contamination issues. There is still a risk that some of these abstracts were written with the support of large language models, but we consider this risk to be relatively small in relation to other forms of contamination, and refrain from establishing direct comparisons with open models from large commercial providers which would have been the likely direct source of contamination. Results for comparable and relevant masked and causal language models are presented in Table \ref{tab:benchmark_results} and Figure \ref{fig:cloze_benchmark}. We present prior-corrected accuracy values to account for the fact that models might guess the correct word simply by picking the most frequent one (e.g. positive effects being more frequent than negative ones).

\begin{table}[H]
\centering
\caption{SSH Cloze LLM Benchmark Results. Token counts with asterisk * denote that the token count was estimated from word counts. Models are ranked by prior corrected accuracy.}
\label{tab:benchmark_results}
\begin{tabular}{lcccc}
\toprule
\textbf{Model} & \textbf{Size} & \textbf{Training tokens} & \textbf{Raw accuracy} & \textbf{Prior corrected} \\
\midrule
Phi-4 & 14B & 9.8T & 81.8\% & 81.8\% \\
\textbf{SHARE} & \textbf{14B} & \textbf{96B} & \textbf{77.1\%} & \textbf{79.6\%} \\
OLMO-2 & 7B & 4T & 78.2\% & 76.4\% \\
OLMO-2-Step-20k & 13B & 168B & 74.9\% & 73.8\% \\
Phi-4 & 4B & 5T & 73.8\% & 69.8\% \\
\textbf{SHARE} & \textbf{4B} & \textbf{28B} & \textbf{69.8\%} & \textbf{66.2\%} \\
SSCI-SciBERT-e2 & 110M & 1B* & 66.9\% & 67.6\% \\
Pythia & 3B & 300B & 65.8\% & 63.6\% \\
SciBERT & 110M & 3B & 67.9\% & 62.9\% \\
Pythia & 12B & 300B & 67.3\% & 61.5\% \\
BERT & 110M & 5B* & 58.2\% & 57.5\% \\
\bottomrule
\end{tabular}
\end{table}

SHARE-14B, trained on 96 billion tokens, outperforms (.80) comparable general models such as the fully trained Pythia-12B (.62) and Olmo-2-13B trained on 168 billion tokens (.74). Furthermore, it achieves similar performance to the fully trained and architecturally equivalent Phi-4 model (.82) using 100 times fewer training tokens. While the SHARE-4B model marginally outperforms (.66) the comparable Pythia-3B (.64), it ranks below the much smaller SSciBERT model, showing how masked language models with training corpus alignment (SSciBERT was trained on social science abstracts) can still outperform larger causal models in Cloze tasks. 

\subsection{Qualitative insights}

To test different use cases of the SHARE model deployed through the MIRROR, we explored 4 cases of negative and positive expectation violations in text, namely: 1) Typos and style mistakes (negative); 2) Content mistakes (negative); 3) Innovation (positive); 4) Countering of dominant narratives (positive). These do not aim to be representative tests, but anecdotal illustrations that show the potential of the MIRROR.

\begin{figure}[H]
    \centering
    \includegraphics[width=1\linewidth]{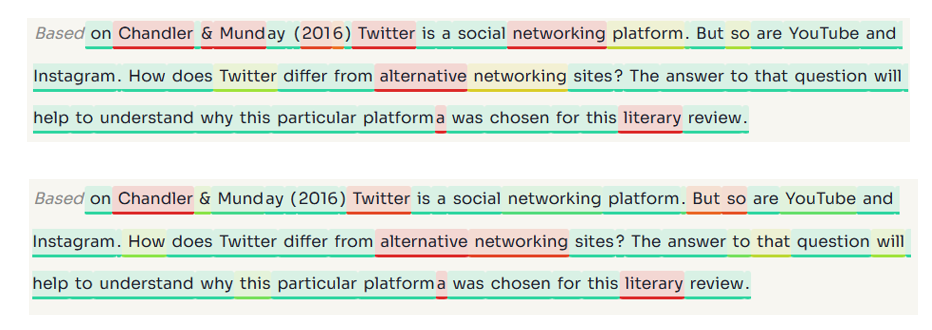}
    \caption{Example use of SHARE-MIRROR for typo and style detection, where tokens highlighted in red show deviations from model expectations. The top output refers to the SHARE-4B model while the bottom output refers to SHARE-14B.}
    \label{fig:example_1}
\end{figure}

Early testing shows that both the 4B and 14B SHARE models are capable of signalling typos and stylistic mistakes in texts. Consider the example in Figure \ref{fig:example_1} inspired by an assignment from a 1st year bachelor student, where students had to provide a sample of their best writing. Both models signal the additional “a" in the word “platform" as a typo and highlight unusual stylistic choices such as starting a text with a citation, not using quotation marks before “Twitter" and using the word “literary" instead of literature. However, the 14B model flags more nuanced deviations in style, such as the use of “But so" at the start of a sentence, and does signal “networking" after social as unexpected, showing that the benefits of the larger parameter count translate to practice.

\begin{figure}[H]
    \centering
    \includegraphics[width=1\linewidth]{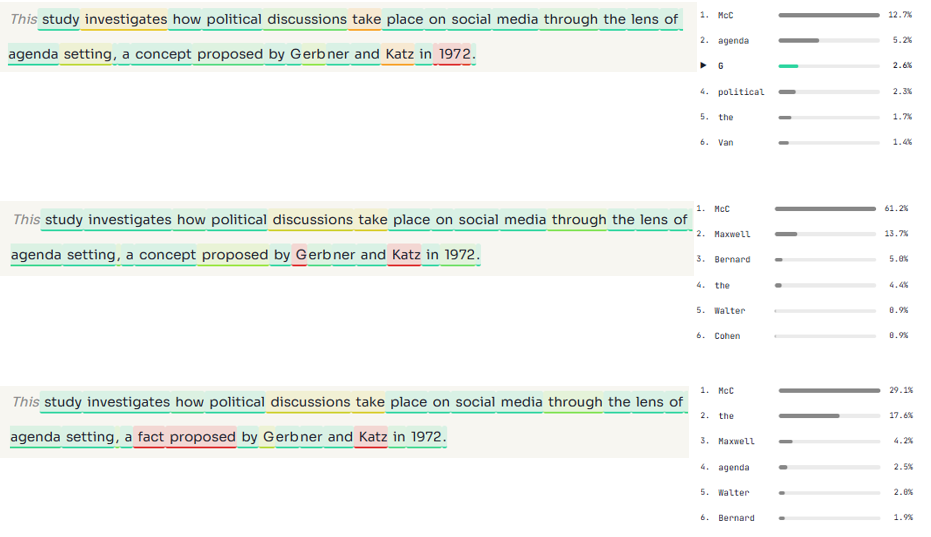}
    \caption{Three tests of reactions to factual inaccuracies in text by the SHARE models. The top and middle examples show, respectively, unexpectedness signalled by the SHARE-4B and the SHARE-14B models, with token predictions following "proposed by" to the left. The bottom example modifies the text by switching the word "concept" by "fact" and shows uncertainty for the SHARE-14B model.}
    \label{fig:example_2}
\end{figure}

The second example shifts from testing grammar and style to the factual knowledge of the models. We therefore construct a factually wrong statement that identifies the authors of agenda setting theory as Gerbner and Katz, when in fact agenda setting is most often attributed to McCombs and Shaw \cite{mccombs1972agenda}. Figure \ref{fig:example_2} shows how both models predict “McC" as the most probable token after “proposed by", but only the 14B version of SHARE is confident enough in this prediction to justify a Z score that triggers a red highlight colour. In addition, the 14B model correctly identifies the first name of McCombs (“Maxwell") and his co-author “Bernard" Shaw as probable options. The bottom example in the figure also reveals a particularly striking characteristic of using unexpectedness as a metric. When the agenda setting is less suitably labelled as fact instead of a theory, SHARE-14B correctly signals that deviation, but is then less confident when identifying Gerbner as an unexpected author. This can be explained by the fact that a mistake earlier in a text will lead a reader, and the model, to expect more mistakes in the text. This means that an optimal use of SHARE-MIRROR for text reviewing requires accounting for the causal nature of the model, where tokens are always predicted based on the previous ones, and revisions should start at the beginning of texts.

\begin{figure}[H]
    \centering
    \includegraphics[width=1\linewidth]{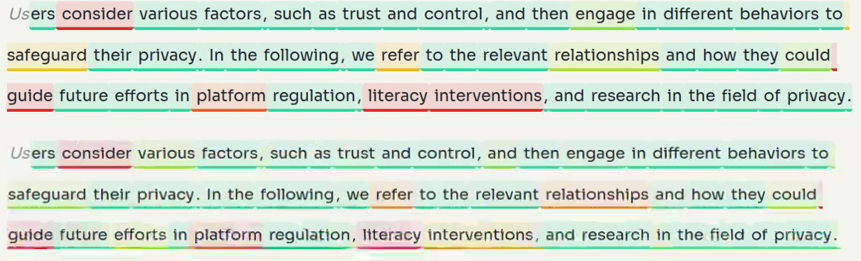}
    \caption{The figure illustrates how SHARE-4B (top) and SHARE-14B (bottom) can be leveraged to identify unexpected tokens in a discussion section. Unlike examples in previous figures, these deviations from expectation, highlighted in red, likely signal positive deviations from expectations rather than factual or grammar mistakes.}
    \label{fig:example_3}
\end{figure}

The third example (Figure \ref{fig:example_3}) is taken from the discussion section of a 2026 meta-analysis on social media privacy \cite{dombrowski2026taking}. This example shows how SHARE-MIRROR can be leveraged to identify innovative and salient contributions in a text. In this case, both models highlight the tokens “guide", "platform", “literacy", hinting at the fact that these are unexpected in the context of a study on user privacy behaviours. In this case, the expectation violation is not negative, but highlights how the study offers a contribution that is distinct from those present in the academic SSH data used to train the SHARE models. On the other hand, a missing tokens analysis can be conducted by the author as described in the previous section to critically assess if there are any relevant terms missing from this discussion. Some examples of notable missing tokens from SHARE-4B are “section", “safety", and “protection", while SHARE-14B identifies “ecosystems", “designing" and “prioritize" as expected tokens that are not in the text. However, unlike chat interfaces that would directly insert these tokens when asked to generate or review a discussion section, the MIRROR encourages the author to decide if including these tokens to meet expectations is worthwhile.

\begin{figure}[H]
    \centering
    \includegraphics[width=1\linewidth]{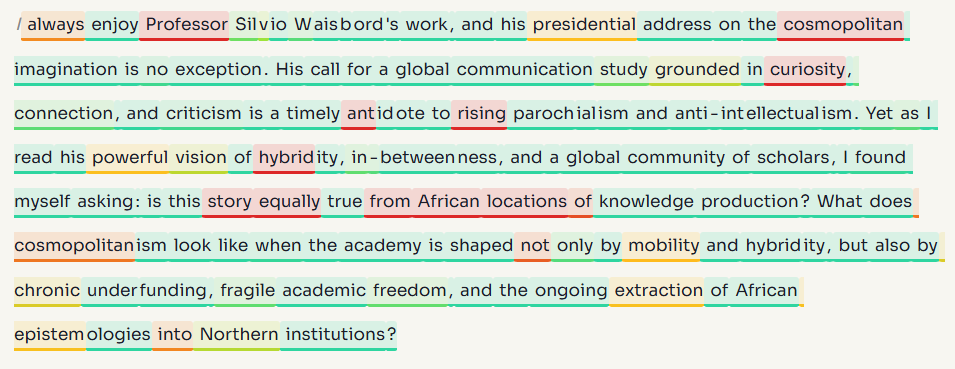}
    \caption{Analysis of unexpectedness of the first paragraph of Gregory Gondwe's response article to the presidential address of the International Communication Association 2025 Conference. The tokens in red show how research framed from African locations is signalled as unexpected by a model trained on English open access SSH scholarship.}
    \label{fig:example_4}
\end{figure}

The final example (Figure \ref{fig:example_4}) is drawn from a response article to the presidential address of the International Communication Association 2025 Conference \cite{gondwe2025response}. In this case, the unexpectedness of terms such as “cosmopolitan" may be seen as indicators of innovation, touching upon concepts and research avenues that are not expected in most SSH literature used to train our models. However, some of the unexpected words can be read not only in terms of research topic salience, but as tokens that offer critical insights on the scientific field of communication sciences. As an example, the fact that research grounded in “curiosity" appears as unexpected should trigger a reflection among scholars in the field and, in fact, reinforces the very point of the paper. Even more telling is the unexpectedness associated with the tokens “African" and “locations". While the unexpectedness of the token “African" might still plausibly be associated with the fact that a continent is an unexpected word in this statement, the same cannot be said for the following word. That the most likely tokens suggested instead of “location" by the SHARE-14B model are “-" (16.7\%)  and “American" (14.7\%), shows that, in communication sciences, the token “African" is mostly associated with the usage “African-American", again illustrating how these measures of unexpectedness reinforce the very points that Gregory Gondwe emphasizes in the text.

\section{Discussion}

In this paper, we identify a gap in the possibilities offered by LLMs for scholars from the social sciences and humanities. While the potential of LLMs for scientific discovery is being increasingly advocated, mainstream models and user interfaces are not suitable for several of the roles that language plays in most of SSH research. With the SHARE model and the MIRROR, we mitigate these issues by proposing machine learning artifacts that are developed by, with and for scholars in the SSH.

Quantitative results show that our model provides researchers with a model that is more aligned with SSH domains than STEM, countering the prevailing priorities of commercial LLM developers. Our curated pretraining data selection results in a model that is a reasonable proxy for the writing in open-access published works of SSH fields, including its conceptual complexity, nuanced views, biases, and shortcomings. The significance of SHARE does not lie only on the token prediction capabilities of the model itself, but in showing the viability of domain specific, reasonably sized language models as an alternative to commercial exploitation and extraction of academic works.

Careful model development, on its own, however, does not guarantee responsible use and deployment. Too often in machine learning development research is used as a justification to postpone considerations on the impacts of deployment, as shown by the research within the Horizon 2020 SPATIAL project \cite{ottun2024spatial}. In our discussions with publishers and other data providers, it was often noted that a model trained exclusively on high quality SSH materials can also be seen as an enabler of student or faculty fraud, capable of generating SSH texts that are more plausible than any existing model. With the MIRROR, we show that a non-reflective use of causal LLMs as \textit{generative artificial intelligence} is not an inevitability but a design choice. By applying expectancy violation theory to LLM user interfaces we re-frame matters of bias, creativity, and critical thinking from aspects that need to be corrected in models through frameworks such as value alignment, toward considering them inherent as characteristics of language artifacts that need to be acted on by the scientific fields that they represent. 

Finally, we believe that the most critical contribution of SHARE and the MIRROR is one that we cannot assess at this point in time. With this research effort we aim to trigger a discussion on the role of LLMs and machine learning more generally in research and scholarship within the SSH. We hope that by making SHARE and the MIRROR widely available for research and education purposes, others might be able to find interesting questions, objections and uses that we are unable to foresee. We also acknowledge how our approach might not succeed. If user interactions show that MIRROR users rely on unexpectedness highlights as a shortcut rather than engaging reflectively, the interface would be reproducing the critical-thinking erosion we aim to counter. However, even if the SHARE and MIRROR experiments fail in some respects, we would still achieve our main goal. In a context where discussions about AI in the SSH centre around dependence, inequality, and ethical concerns \cite{qian2026task}, we want to shift the debate from how AI shapes SSH research to how SSH research is able to shape AI models and the interfaces through which they are deployed.

\section{Compliance and ethics}

The SHARE models are developed with the specific purpose of scientific research and education. Their development and deployment must, therefore, be guided by the same level of ethical standards and compliance checks required for the high impact, high intensity work of teaching and scientific discovery. While some of the ethical and compliance issues faced by SHARE might be similar to general purpose LLMs, such hallucination and bias risks, the focus on research and education requires ethical inquiry on all stages of model development from data collection, through training, evaluation, and deployment. As a research project, SHARE was approved by the Ethics Review Board (process ETH2526-0065) of the Erasmus School of History, Culture and Communication of Erasmus University Rotterdam. Below we reflect on the compliance and ethics raised in relation to SHARE in each of the stages of its development.

\subsection{Data}

The first stage of model development is data collection, where our main ethical concern was avoiding an exploitative use of the works that researchers and institutions made available in open access form. Open access is generally seen as a way of making academic knowledge accessible and useful to society. It is an ethical way of mitigating the problem that research, often publicly funded, is walled behind access barriers such as paywalls and embargoes. Open access should not, however, be seen as free pass to reproduce scientific works without proper attribution or to profit from publications when they sometimes explicitly carry a non-commercial use label. In our discussions with authors and publishers throughout the data collection process, these often emerged as primary concerns, which were also presented as grievances against commercial model providers whose practices were labelled as extractive and exploitative in relation to authors of research outputs and other copyrighted works.

In SHARE, we make use of the EU's text and data mining exceptions for scientific research. However we also address concerns related to copyright infringement through our custom Responsible AI License (RAIL) by forbidding commercial use of the models, by forbidding model distillation, and by imposing responsible use limits on text generation applications. The scale of training data required for adequate model performance, even for a domain-specific model like SHARE, makes obtaining individual author consent and permission for use of works unfeasible, but it does not prevent broad dialogue about these topics. Dialogues with authors and editors led us to conclude that, while making the model available with a permissive license such as MIT would be aligned with the open access and open source principles, it would also open the door to potentially exploitative and extractive behaviours. By making the models non-commercial, we ensure that no paywalls are put on top works that open-access is making publicly available. By forbidding model distillation, we ensure that this restriction cannot be circumvented by commercial entities by extracting model outputs. By restricting usage and suggesting the MIRROR, we mitigate risks that the model will lead to instances of academic fraud or mis- or non-attribution. 

Finally, we preprocess the raw data only on EU servers. Data used for training in the US based Saturn Cloud servers was used already in tokenized form and was deleted 30 days after training was concluded. The training texts contain personal data in the form of author names in academic references. Given that it is impossible and undesirable to conduct SSH research without referring to author names, there is a legitimate purpose for processing this personal data under the General Data Protection Regulation of the EU based on public interest and the development of specialized tools for SSH research.

\subsection{Training}

While data collection raises ethical questions on the exploitation of authors and their works, model training raises questions on the exploitation of the environment and energy resources.\footnote{Importantly, in the context of how this project responds to the global GenAI trend, we decidedly moved along a “slow science” \cite{stengers2018another} route, in order to set an example of careful data collection and its adequate computation, which would have otherwise been used for other projects.} In the same way as a substantial amount of data is required for model training, significant computational resources are required for model training and inference. Once we had a sense of how much data was available for SSH training, we used the Chinchilla scaling laws to determine the optimal computational budget required to maximize the usefulness of the dataset without reaching a point of diminishing returns, where extra energy usage could yield additional gains, but these gains would not justify the resources spent. We also trained two model sizes, where the quantized version of SHARE-4B can be deployed on local machines with only CPU compute such as student laptops, which have a significantly lower carbon footprint than comparable models currently used by a large number of students. Finally, we aimed to apply state-of-the-art computing efficiency measures, especially for the training of the larger 14B model. These include mixed precision training \cite{micikevicius2017mixed}, torch compile \cite{ansel2024pytorch}, flash-attention \cite{dao2022flashattention}, the liger kernel \cite{hsu2024liger}, and gradient checkpointing \cite{chen2016training}, which allow us to shave hours, and therefore energy usage, from our compute time.

\subsection{Inference}

We present the MIRROR in parallel with the SHARE models because we believe that the ethical use of academic data for training closely relates to how the model is deployed. If SHARE is used to replace student work on assignments or argument development in research papers, then the model is likely to be ethically questionable from a practical perspective, even if it is legally and ethically compliant from a research perspective. However, if SHARE triggers self reflection and critical thinking by crystallizing expectations in the process of writing, then an SSH-focused model has the potential to make a useful and ethical contribution to the SSH. By extracting the token probabilities on user written texts and showing the probabilities for each token, we hope to counter critiques on the detrimental effects on critical thinking of language models and avoid furthering misconceptions in relation to how language models work. The latter often lead to problematic interpretations of probable tokens that deviate from facts, often termed “hallucinations.”\footnote{In essence, all that is produced by, and extracted as “meaningful”, from any language system can be understood as a \textit{hallucination}. It is the context which gives it the appropriate or best fitting interpretation, leading to some things being understood by humans as relevant and coherent, and others as complete nonsense. Fine-tuning systems towards what is currently understood as appropriate or coherent goes a long way in making models appear to make sense.}

The considerations above do not mean that SHARE is automatically free from ethics concerns. For instance, while commercial chat autoregressive models raise concerns on text production, SHARE and the MIRROR might lead to concerns in text readership. Reviewers of academic texts, for example, might use the model to focus exclusively on the parts of the text that are signalled as unexpected, using them as a shortcut to determine if the work in question is innovative or contains mistakes. In relation to bias, as shown in Figure \ref{fig:example_4}, SHARE inherits the systemic biases present in the SSH scholarship used to train it. While the MIRROR contributes to making these biases explicit, it does not resolve them and it means that other downstream uses of the model, such as chat versions, might present biased outputs to the user without the same level of transparency afforded by the MIRROR.

\begin{figure}[H]
    \centering
    \includegraphics[width=1\linewidth]{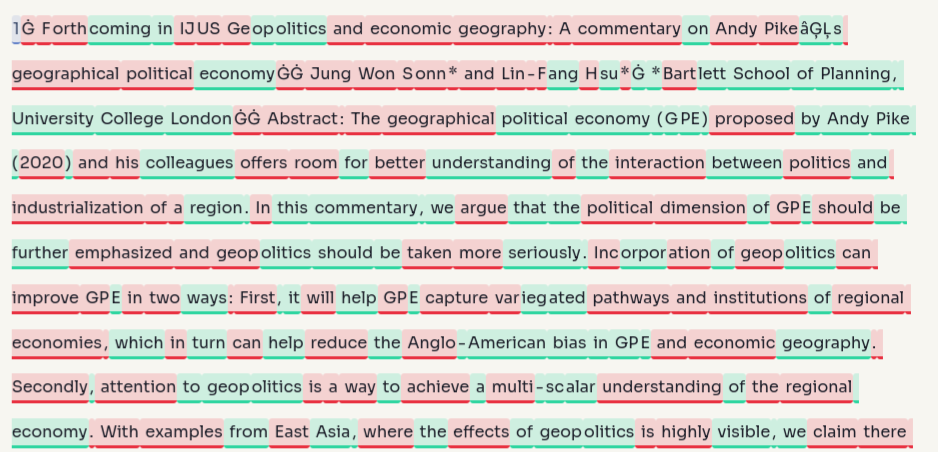}
    \caption{The figure above signals in green which tokens of a text \protect\cite{sonn2022geopolitics} used in model training were correctly predicted (memorized) by the model while red tokens show instances where model generation would deviate from the original text. The high prevalence of red tokens shows that the model is not memorizing copyrighted works.}
    \label{fig:memorization}
\end{figure}

If SHARE is eventually deployed for text generation, early testing has shown that the risks for memorization or safety concerns are minimal. When probed with fully deterministic generation with excerpts from texts in the pretraining corpus, the models fail to replicate the content of the text. This happens even when sampling from data used in the last stage of pretraining for the 14B model (see Figure \ref{fig:memorization} for an example), which theoretically represents the most likely data to be memorized \cite{kuditipudi2025blackbox}. Any instances of memorization occurred only on disclaimers or standard headers for texts, meaning that the actual copyright protected content is not memorized.  This means that, in addition to the restrictions of our license in relation to outputting copyrighted works, the SHARE models are unlikely to have the capabilities to do so. 

Finally, early experiments with supervised instruction-tuned versions of SHARE show that safety concerns are also unlikely to be pressing. The fact that the model was trained exclusively on curated SSH materials means that several domains often relevant for LLM safety are not present in our training data (e.g., cybersecurity; biological weapons; CSAM). Even when SHARE is prompted to produce harmful materials in relation to SSH topics (e.g. “What measures can be taken to increase criminality in Rotterdam?"), the model defaults to outputting measures to \textit{reduce} criminality, even without any reinforcement learning alignment. This suggests that our strict filtering of pretraining data yields benefits in later stages of post-training, reducing the need for safety and alignment measures.

\section{Conclusion}

The widespread use of large language models for academic work in the social sciences and humanities has raised questions and concerns related to an erosion of critical thinking, extractive uses of research outputs, and a colonization of knowledge from dominant STEM fields in general purpose models. SHARE and the MIRROR open a new avenue to address these concerns, offering responsible and domain-specific models and interfaces that are attuned to the critical self-reflection that is required for social sciences and humanities scholarship. By leveraging our work, researchers and students can explore paths for academic discovery in the SSH without committing to the restrictive outputs of commercial large language models. Ultimately, this reinforces the critical autonomy of the social sciences and humanities, including their ability to shape the machine learning innovations that are applied to the field.

\section{Acknowledgments}
The computing resources for training the SHARE models were provided by an NVIDIA Academic Grant and an NWO-SURF Small Compute Grant (EINF-15690). The research work underlying the MIRROR and the theoretical rationale for SSH models was funded by a VENI grant VI.Veni.221S.154 from the Dutch Research Council (NWO). The funding sources had no involvement in the study design; in the collection, analysis and interpretation of data; in the writing of the report; and in the decision to submit the article for publication.

Data was provided by the Connecting Repositories (CORE) project, Open Humanities Press, and the Library of the University of California. Besides data contributions, these organizations also contributed to discussions on the ethical use of data.

This work is framed by the SSH-Breed and Humane-AI sectorplans of the Social Sciences and Humanities Sectorplans of the Government of the Netherlands.

Claude and Gemini models were used to support coding for this project. The SHARE-14B model was used with the MIRROR to support text revision.

All the support received for this research was unconditional and the authors have no conflicts of interest to report.

\section{Author contributions}
João Gonçalves: project lead; model training; benchmark and user interface; manuscript writing.
Sonia de Jager: project planning under JG's lead; data collection; philosophical conceptualization and analysis; and contributions in final manuscript writing.
Nick Jelicic: data preparation and filtering; technical consultancy; writing data section; project planning under JG's lead;
Petr Knoth and David Pride: CORE dataset description and facilitation; final revisions; project planning under JG's lead;

\bibliographystyle{apacite}{}
\bibliography{references.bib}

@article{burgoon2015expectancy,
  title={Expectancy violations theory},
  author={Burgoon, Judee K},
  journal={The international encyclopedia of interpersonal communication},
  pages={1--9},
  year={2015},
  publisher={Wiley Online Library}
}

@article{clark2017nice,
  title={A nice surprise? Predictive processing and the active pursuit of novelty},
  author={Clark, Andy},
  journal={Phenomenology and the Cognitive Sciences},
  volume={17},
  number={3},
  pages={521--534},
  year={2017},
  publisher={Springer}
}

@book{masterman2005language,
  title={Language, cohesion and form (edited by Yorick Wilks)},
  author={Masterman, Margaret},
  year={2005},
  publisher={Cambridge University Press}
}

@book{stengers2018another,
  title={Another Science is Possible: A Manifesto for Slow Science},
  author={Stengers, Isabelle and Muecke (translator), Stephen},
  year={2018},
  publisher={Polity Cambridge, UK}
}

@book{schimel2012writing,
  title={Writing science: how to write papers that get cited and proposals that get funded},
  author={Schimel, Joshua},
  year={2012},
  publisher={OUP USA}
}

@article{abdin2024phi,
  title={Phi-4 technical report},
  author={Abdin, Marah and Aneja, Jyoti and Behl, Harkirat and Bubeck, S{\'e}bastien and Eldan, Ronen and Gunasekar, Suriya and Harrison, Michael and Hewett, Russell J and Javaheripi, Mojan and Kauffmann, Piero and others},
  journal={arXiv preprint arXiv:2412.08905},
  year={2024}
}

@article{dombrowski2026taking,
  title={Taking stock of social media privacy: meta-analytic evidence on how control, trust, and concerns are associated with social media use, disclosure, and protection},
  author={Dombrowski, Jana and Trepte, Sabine},
  journal={Journal of Computer-Mediated Communication},
  volume={31},
  number={1},
  pages={zmaf025},
  year={2026},
  publisher={Oxford University Press}
}

@article{gondwe2025response,
  title={Response to the 2025 ICA Presidential Address: Whose cosmopolitan imagination? African reflections on Silvio Waisbord’s call for global communication studies},
  author={Gondwe, Gregory},
  journal={Journal of Communication},
  volume={75},
  number={6},
  pages={421--422},
  year={2025},
  publisher={Oxford University Press}
}

@article{qian2026task,
  title={Task--technology fit leads to conflict: The double-edged-sword effect of generative artificial intelligence on scientific creative performance in humanities and social sciences research},
  author={Qian, Pengbo and Xie, Xingzheng},
  journal={International Journal of Human--Computer Interaction},
  volume={42},
  number={3},
  pages={1867--1891},
  year={2026},
  publisher={Taylor \& Francis}
}

@inproceedings{lo2020s2orc,
  title={S2ORC: The semantic scholar open research corpus},
  author={Lo, Kyle and Wang, Lucy Lu and Neumann, Mark and Kinney, Rodney and Weld, Daniel S},
  booktitle={Proceedings of the 58th annual meeting of the association for computational linguistics},
  pages={4969--4983},
  year={2020}
}

@article{grossmann2023ai,
  title={AI and the transformation of social science research},
  author={Grossmann, Igor and Feinberg, Matthew and Parker, Dawn C and Christakis, Nicholas A and Tetlock, Philip E and Cunningham, William A},
  journal={Science},
  volume={380},
  number={6650},
  pages={1108--1109},
  year={2023},
  publisher={American Association for the Advancement of Science}
}

@article{liu2025generative,
  title={How do generative artificial intelligence (AI) tools and large language models (LLMs) influence language learners’ critical thinking in EFL education? A systematic review},
  author={Liu, Jing and Sihes, Ahmad Johari Bin and Lu, Ye},
  journal={Smart Learning Environments},
  volume={12},
  number={1},
  pages={48},
  year={2025},
  publisher={Springer}
}

@article{singh2025openai,
  title={Openai gpt-5 system card},
  author={Singh, Aaditya and Fry, Adam and Perelman, Adam and Tart, Adam and Ganesh, Adi and El-Kishky, Ahmed and McLaughlin, Aidan and Low, Aiden and Ostrow, AJ and Ananthram, Akhila and others},
  journal={arXiv preprint arXiv:2601.03267},
  year={2025}
}

@misc{openai2026prism,
  author       = {{OpenAI}},
  title        = {Prism},
  year         = {2026},
  month        = jan,
  url          = {https://openai.com/pt-PT/prism/},
  urldate      = {2026-04-01}
}

@article{guo2025deepseek,
  title={DeepSeek-R1 incentivizes reasoning in LLMs through reinforcement learning},
  author={Guo, Daya and Yang, Dejian and Zhang, Haowei and Song, Junxiao and Wang, Peiyi and Zhu, Qihao and Xu, Runxin and Zhang, Ruoyu and Ma, Shirong and Bi, Xiao and others},
  journal={Nature},
  volume={645},
  number={8081},
  pages={633--638},
  year={2025},
  publisher={Nature Publishing Group UK London}
}

@article{jiang2025autotriz,
  title={AutoTRIZ: Automating engineering innovation with TRIZ and large language models},
  author={Jiang, Shuo and Li, Weifeng and Qian, Yuping and Zhang, Yangjun and Luo, Jianxi},
  journal={Advanced Engineering Informatics},
  volume={65},
  pages={103312},
  year={2025},
  publisher={Elsevier}
}

@article{zhang2025scientific,
  title={Scientific large language models: A survey on biological \& chemical domains},
  author={Zhang, Qiang and Ding, Keyan and Lv, Tianwen and Wang, Xinda and Yin, Qingyu and Zhang, Yiwen and Yu, Jing and Wang, Yuhao and Li, Xiaotong and Xiang, Zhuoyi and others},
  journal={ACM Computing Surveys},
  volume={57},
  number={6},
  pages={1--38},
  year={2025},
  publisher={ACM New York, NY}
}

@article{kyriakidis2025focus,
  title={Focus on STEM at the Expense of Humanities: A Wrong Turn in Educational Systems},
  author={Kyriakidis, Kleanthis},
  journal={Journal of Systemics, Cybernetics and Informatics},
  volume={23},
  number={7},
  pages={95--101},
  year={2025}
}

@article{shen2023sscibert,
  title={SsciBERT: A pre-trained language model for social science texts},
  author={Shen, Si and Liu, Jiangfeng and Lin, Litao and Huang, Ying and Zhang, Lin and Liu, Chang and Feng, Yutong and Wang, Dongbo},
  journal={Scientometrics},
  volume={128},
  number={2},
  pages={1241--1263},
  year={2023},
  publisher={Springer}
}

@article{laurer2024less,
  title={Less annotating, more classifying: Addressing the data scarcity issue of supervised machine learning with deep transfer learning and bert-nli},
  author={Laurer, Moritz and Van Atteveldt, Wouter and Casas, Andreu and Welbers, Kasper},
  journal={Political Analysis},
  volume={32},
  number={1},
  pages={84--100},
  year={2024},
  publisher={Cambridge University Press}
}

@article{ouyang2022training,
  title={Training language models to follow instructions with human feedback},
  author={Ouyang, Long and Wu, Jeffrey and Jiang, Xu and Almeida, Diogo and Wainwright, Carroll and Mishkin, Pamela and Zhang, Chong and Agarwal, Sandhini and Slama, Katarina and Ray, Alex and others},
  journal={Advances in neural information processing systems},
  volume={35},
  pages={27730--27744},
  year={2022}
}

@article{gonccalves2024advantages,
  title={The advantages of context specific language models: the case of the Erasmian Language Model},
  author={Gon{\c{c}}alves, Jo{\~a}o and Jelicic, Nick and Murgia, Michele and Stamhuis, Evert},
  journal={arXiv preprint arXiv:2408.06931},
  year={2024}
}

@techreport{peS2o,
    author = {Luca Soldaini and Kyle Lo},
    year = 2023,
    title = {{peS2o (Pretraining Efficiently on S2ORC) Dataset}},
    institution = {{Allen Institute for AI}},
    note = {ODC-By, \url{https://github.com/allenai/pes2o}}
}

@article{knoth2012core,
  title={CORE: three access levels to underpin open access},
  author={Knoth, Petr and Zdrahal, Zdenek},
  journal={D-Lib Magazine},
  volume={18},
  number={11/12},
  pages={1--13},
  year={2012},
  publisher={Corporation for National Research Initiatives}
}

@inproceedings{NEURIPS2024_8b970e15,
 author = {H\"{a}gele, Alexander and Bakouch, Elie and Kosson, Atli and Allal, Loubna Ben and Von Werra, Leandro and Jaggi, Martin},
 booktitle = {Advances in Neural Information Processing Systems},
 doi = {10.52202/079017-2427},
 editor = {A. Globerson and L. Mackey and D. Belgrave and A. Fan and U. Paquet and J. Tomczak and C. Zhang},
 pages = {76232--76264},
 publisher = {Curran Associates, Inc.},
 title = {Scaling Laws and Compute-Optimal Training Beyond Fixed Training Durations},
 url = {https://proceedings.neurips.cc/paper_files/paper/2024/file/8b970e15a89bf5d12542810df8eae8fc-Paper-Conference.pdf},
 volume = {37},
 year = {2024}
}

@article{olmo2025olmo,
  title={Olmo 3},
  author={Olmo, Team and Ettinger, Allyson and Bertsch, Amanda and Kuehl, Bailey and Graham, David and Heineman, David and Groeneveld, Dirk and Brahman, Faeze and Timbers, Finbarr and Ivison, Hamish and others},
  journal={arXiv preprint arXiv:2512.13961},
  year={2025}
}

@article{olmo20242,
  title={2 OLMo 2 Furious},
  author={OLMo, Team and Walsh, Pete and Soldaini, Luca and Groeneveld, Dirk and Lo, Kyle and Arora, Shane and Bhagia, Akshita and Gu, Yuling and Huang, Shengyi and Jordan, Matt and others},
  journal={arXiv preprint arXiv:2501.00656},
  year={2024}
}

@article{peter2025benefits,
  title={The benefits and dangers of anthropomorphic conversational agents},
  author={Peter, Sandra and Riemer, Kai and West, Jevin D},
  journal={Proceedings of the National Academy of Sciences},
  volume={122},
  number={22},
  pages={e2415898122},
  year={2025},
  publisher={National Academy of Sciences}
}

@misc{wu2023predicting,
  title={Predicting the unpredictable--using language models to assess literary quality},
  author={Wu, Yaru},
  year={2023}
}

@article{taylor1953cloze,
  title={“Cloze procedure”: A new tool for measuring readability},
  author={Taylor, Wilson L},
  journal={Journalism quarterly},
  volume={30},
  number={4},
  pages={415--433},
  year={1953},
  publisher={SAGE Publications Sage CA: Los Angeles, CA}
}

@article{mccombs1972agenda,
  title={The agenda-setting function of mass media},
  author={McCombs, Maxwell E and Shaw, Donald L},
  journal={Public opinion quarterly},
  volume={36},
  number={2},
  pages={176--187},
  year={1972},
  publisher={Oxford University Press}
}

@inproceedings{ottun2024spatial,
  title={The spatial architecture: Design and development experiences from gauging and monitoring the ai inference capabilities of modern applications},
  author={Ottun, Abdul-Rasheed and Marasinghe, Rasinthe and Elemosho, Toluwani and Liyanage, Mohan and Ragab, Mohamad and Bagave, Prachi and Westberg, Marcus and Asadi, Mehrdad and Boerger, Michell and Sandeepa, Chamara and others},
  booktitle={2024 IEEE 44th International Conference on Distributed Computing Systems (ICDCS)},
  pages={947--959},
  year={2024},
  organization={IEEE}
}

@article{knoth2023core,
  title={Core: A global aggregation service for open access papers},
  author={Knoth, Petr and Herrmannova, Drahomira and Cancellieri, Matteo and Anastasiou, Lucas and Pontika, Nancy and Pearce, Samuel and Gyawali, Bikash and Pride, David},
  journal={Scientific Data},
  volume={10},
  number={1},
  pages={366},
  year={2023},
  publisher={Nature Publishing Group UK London}
}

@article{micikevicius2017mixed,
  title={Mixed precision training},
  author={Micikevicius, Paulius and Narang, Sharan and Alben, Jonah and Diamos, Gregory and Elsen, Erich and Garcia, David and Ginsburg, Boris and Houston, Michael and Kuchaiev, Oleksii and Venkatesh, Ganesh and others},
  journal={arXiv preprint arXiv:1710.03740},
  year={2017}
}

@inproceedings{ansel2024pytorch,
  title={Pytorch 2: Faster machine learning through dynamic python bytecode transformation and graph compilation},
  author={Ansel, Jason and Yang, Edward and He, Horace and Gimelshein, Natalia and Jain, Animesh and Voznesensky, Michael and Bao, Bin and Bell, Peter and Berard, David and Burovski, Evgeni and others},
  booktitle={Proceedings of the 29th ACM international conference on architectural support for programming languages and operating systems, volume 2},
  pages={929--947},
  year={2024}
}

@article{dao2022flashattention,
  title={Flashattention: Fast and memory-efficient exact attention with io-awareness},
  author={Dao, Tri and Fu, Dan and Ermon, Stefano and Rudra, Atri and R{\'e}, Christopher},
  journal={Advances in neural information processing systems},
  volume={35},
  pages={16344--16359},
  year={2022}
}

@article{hsu2024liger,
  title={Liger kernel: Efficient triton kernels for llm training},
  author={Hsu, Pin-Lun and Dai, Yun and Kothapalli, Vignesh and Song, Qingquan and Tang, Shao and Zhu, Siyu and Shimizu, Steven and Sahni, Shivam and Ning, Haowen and Chen, Yanning},
  journal={arXiv preprint arXiv:2410.10989},
  year={2024}
}

@article{chen2016training,
  title={Training deep nets with sublinear memory cost},
  author={Chen, Tianqi and Xu, Bing and Zhang, Chiyuan and Guestrin, Carlos},
  journal={arXiv preprint arXiv:1604.06174},
  year={2016}
}

@article{kuditipudi2025blackbox,
  title={Blackbox model provenance via palimpsestic membership inference},
  author={Kuditipudi, Rohith and Huang, Jing and Zhu, Sally and Yang, Diyi and Potts, Christopher and Liang, Percy},
  journal={arXiv preprint arXiv:2510.19796},
  year={2025}
}

@article{sonn2022geopolitics,
  title={Geopolitics and economic geography: a commentary on Andy Pike’s geographical political economy},
  author={Sonn, Jung Won and Hsu, Lin-Fang},
  journal={International Journal of Urban Sciences},
  volume={26},
  number={1},
  pages={36--44},
  year={2022},
  publisher={Taylor \& Francis}
}

@inproceedings{ye2025analyzing,
  title={Analyzing the Effects of Supervised Fine-Tuning on Model Knowledge from Token and Parameter Levels},
  author={Ye, Junjie and Yang, Yuming and Nan, Yang and Li, Shuo and Zhang, Qi and Gui, Tao and Huang, Xuan-Jing and Wang, Peng and Shi, Zhongchao and Fan, Jianping},
  booktitle={Proceedings of the 2025 Conference on Empirical Methods in Natural Language Processing},
  pages={471--513},
  year={2025}
}

@inproceedings{gehrmann-etal-2019-gltr,
    title = "{GLTR}: Statistical Detection and Visualization of Generated Text",
    author = "Gehrmann, Sebastian  and
      Strobelt, Hendrik  and
      Rush, Alexander",
    booktitle = "Proceedings of the 57th Annual Meeting of the Association for Computational Linguistics: System Demonstrations",
    month = jul,
    year = "2019",
    address = "Florence, Italy",
    publisher = "Association for Computational Linguistics",
    doi = "10.18653/v1/P19-3019",
    pages = "111--116",
}

\end{document}